\documentclass[runningheads]{llncs}
\usepackage{graphicx}
\usepackage{amsmath,amssymb} 
\usepackage{color}
\usepackage{multirow}
\usepackage[width=122mm,left=12mm,paperwidth=146mm,height=193mm,top=12mm,paperheight=217mm]{geometry}
\newcommand{\hide}[1]{}
\newcommand{\ie}[0]{i.e. }
\newcommand{\eg}[0]{e.g. }
\newcommand{\vs}[0]{vs. }

\renewcommand{\vec}[1]{\mathbf{#1}}
\newcommand{\mat}[1]{\mathbf{#1}}
\begin{document}
\pagestyle{headings}
\mainmatter
\def\ECCV18SubNumber{2173}  

\title{Verification of Very Low-Resolution Faces \\ Using An Identity-Preserving Deep Face Super-resolution Network} 

\titlerunning{Verification of Very LR Faces}

\authorrunning{Cansizoglu, Jones, Zhang, Sullivan}

\author{Esra Ataer-Cansizoglu, Michael Jones, Ziming Zhang and Alan Sullivan}
\institute{Mitsubishi Electric Research Labs (MERL)\\Cambridge, MA USA}

\maketitle

\begin{abstract}
Face super-resolution methods usually aim at producing visually
appealing results rather than preserving distinctive features for
further face identification. In this work, we propose a deep learning
method for face verification on very low-resolution face images that
involves identity-preserving face super-resolution. Our framework
includes a super-resolution network and a feature extraction
network. We train a VGG-based deep face recognition
network~\cite{Parkhi2015} to be used as feature extractor. Our
super-resolution network is trained to minimize the feature distance
between the high resolution ground truth image and the super-resolved
image, where features are extracted using our pre-trained feature
extraction network. We carry out experiments on FRGC, Multi-PIE,
LFW-a, and MegaFace datasets to evaluate our method in controlled and
uncontrolled settings. The results show that the presented method
outperforms conventional super-resolution methods in low-resolution
face verification.
\end{abstract}

\section{Introduction}

Face images appear in various platforms and are vital for many applications ranging from forensics to health monitoring.  In most cases, these images are in low-resolution, making face identification difficult. Although many algorithms have been developed for face recognition from high-quality images, few studies focus on the problem of very low-resolution face recognition\hide{ where faces appear smaller than $32\times24$ pixels~\cite{Wang2014}}. The performance of the traditional face recognition algorithms developed for high quality images, degrades considerably on low-resolution faces. 


There exists a tremendous amount of work in image enhancement and upsampling. Recently, high magnification factors greater than $4$ times have gained more attention for targeted objects such as faces with the rise in deep learning methods. Existing\hide{ techniques are designed to  minimize the intensity difference between high-resolution and super-resolved image pairs. As a result these} methods provide an upsampling of the image that is as close  as possible to ``{\em a} face image". Since resulting upsampled images are meant to be used in face identification task, recovering ``{\em the} face" is essential. We present a face super-resolution method that preserves the identity of the person during super-resolution by minimizing the distance in feature space as opposed to the traditional face super-resolution methods designed to minimize the distance in high-resolution image space. 

The goal of this paper is to verify whether a given low-resolution face image is the same person as in a high-resolution gallery image. Our focus is very low-resolution face images with a tiny visible facial area (as low as $6\times6$ pixels). \hide{We consider a scaling factor of $8$, where low-resolution images have a facial size of $8\times8$ pixels. }We represent a face image with its VGG face descriptor. Since our goal is verification, the face descriptor of a super-resolved face image should be as close as possible to the face descriptor of its ground truth high-resolution version. Thus, we train a super-resolution network by minimizing the feature distance between them. Contrary to the conventional face hallucination methods, we consider face descriptor similarity instead of appearance similarity during super-resolution. Moreover, we also perform detailed experiments in order to investigate the effect of various losses in training super-resolution for the task of low-resolution face verification.

\begin{figure}[t]
\centering
\includegraphics[width=0.85\textwidth]{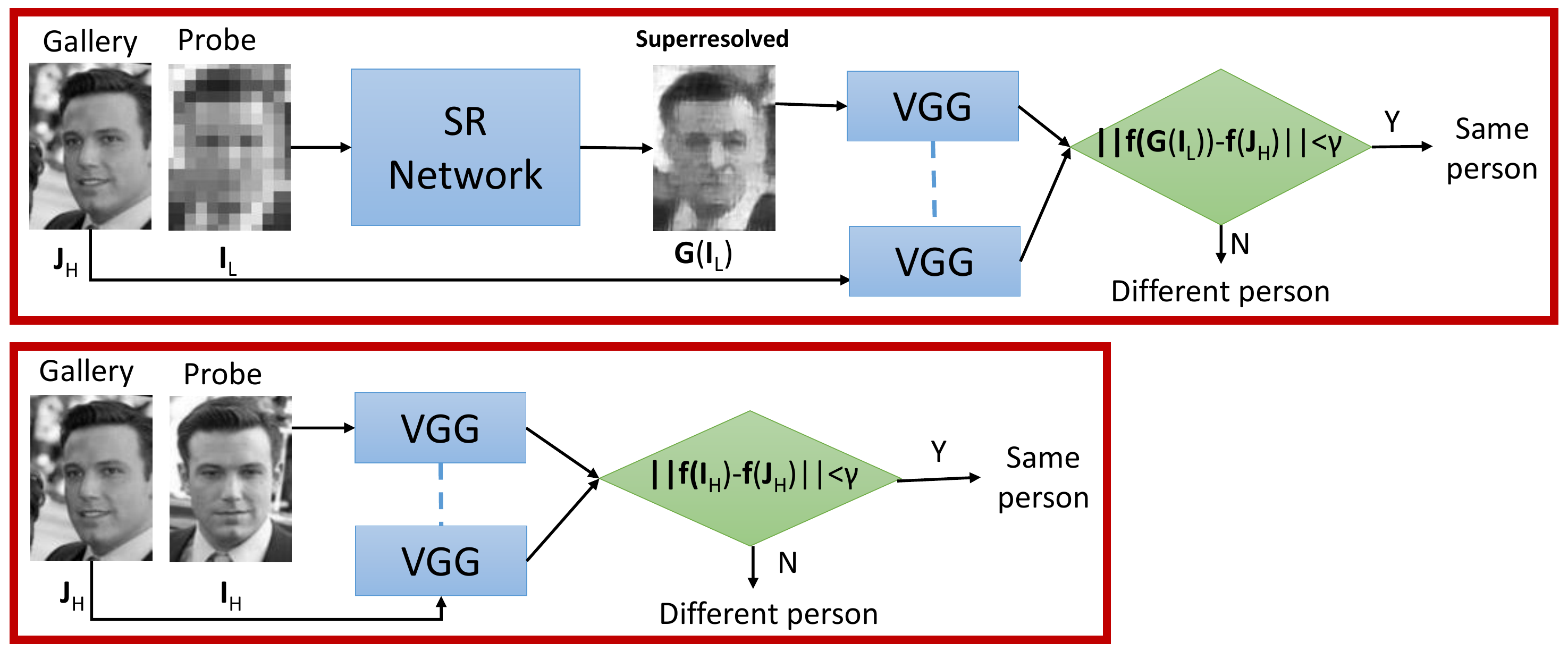}
\caption{System overview for (top) low-resolution and (bottom) high-resolution-resolution face verification. Dashed lines indicate weight sharing between networks.}
\label{fig:overview}
\end{figure}

The performance of super-resolution methods are evaluated by using image quality assessment measures such as peak signal-to-noise ratio (PSNR)\hide{ and structural similarity (SSIM)}. These measures account for the visual similarity of two images by equally paying attention to every pixel in intensity domain. However, face identification relies on discriminative features.\hide{ In order to assess how much a super-resolution method distorts the identity of the face, we should consider the similarity of the images in feature domain.} In this work, we present face descriptor similarity as an evaluation measure to assess the capacity of a method in preserving identity.

The main contributions of our study include: (1) a novel loss term to be used for face super-resolution in order to preserve identity for aggressive scaling factors as big as $8$, (2) an evaluation measure to account for identity preservation on the super-resolved faces, and (3) a thorough analysis of various loss terms in training a face super-resolution network for low-resolution face verification.

\hide{
Face upsampling/SR is important for many applications, espcially high-resolution order magnification (>4x)

Face hallucination 
tries to construct visually appealing faces, but does it provide "the face"

We propose a face super resolution technique that is designed to preserve identity.

Contributions:
1) A novel loss term to be used for face hallucination in order to preserve identity
2) A novel evaluation measure on the super resolved faces
3) Through analysis of various loss terms in training face SR network

What?
Face verification in low-resolution images

How
Using a face super-resolution method

Why it works and why is it different:
Our face SR method aims at preserving identity -- we minimize a loss term that minimizes the distance between superresolved faces and ground truth HR faces
The us eof recognition as a constraint in SR is not new, but using a deep network (VGG) for this constraint is new
Other SR-based face verifictaion methods care about appearance of the SR image, while in our work we only care about verification performance. (Is this true that we are the only ones who is using a single recognition loss)

faces are everywhere
in most important applications, they are in low-res
there are many studies for high-res fr, but few focus on lr fr.

}

\subsection{Related Work}

To solve the problem of low-resolution face verification (in which a
low-resolution probe face is compared to a high-resolution gallery
face) the main problem is how to handle the mismatch in resolutions.
There are two basic approaches to solve this problem.  The first is to
map the low-resolution probe face and the high-resolution gallery face
to a common feature space.  Methods such as coupled locality
preserving mappings~\cite{LiEtAl2010}, coupled kernel
embeddings~\cite{RenEtAl2012}, and multidimensional
scaling~\cite{BiswasEtAl2013} follow this approach.  Unfortunately, finding a resolution-robust feature space is very hard especially for high magnification factors.  The second
approach is to upsample the low-resolution face image (using a
super-resolution algorithm) and compare to the high-resolution
gallery face using a standard face recognition method.  Our method
falls into the second category.

Baker and Kanade~\cite{BakerKanade2000} showed how to greatly improve
super-resolution quality specifically for faces using pairs of high
and low-resolution examples of faces.  Since their work there have
been many papers on face-specific techniques for
super-resolution~\cite{Wang2014survey,BakerKanade2002,LiuEtAl2001,CapelZisserman2001,LiLin2004,LiuEtAl2007}.
All of these methods try to produce visually pleasing high-resolution
face images given the low-resolution input and face-specific models.
They are not directly concerned with improving face recognition
accuracy on the upsampled faces.  However, since the main application
of face super-resolution is face recognition, it makes sense to
optimize a face-specific super-resolution algorithm explicitly to
improve face recognition accuracy.  This idea has been explored in a
number of
papers~\cite{JiaGong2005,Hennings-YeomansEtAl2008,ZouYuen2012}.  The
basic idea is to find a high-resolution face image that optimizes both
reconstruction and recognition costs simultaneously.  They mainly use
linear models for extracting face recognition feature vectors
(e.g. PCA and LDA).  These papers were all written before the recent
era of deep neural networks which now dominate the face recognition
field because of their high accuracy.  The older
recognition-optimizing face-specific super-resolution algorithms work
well for frontal faces taken in controlled environments, but do not
work nearly as well in typical ``in-the-wild'' settings for which deep
networks are so effective.

Recently, a few papers have used convolutional neural networks (CNNs) or generative adversarial
networks (GANs) for face-specific super-resolution.  These methods
better handle uncontrolled input faces with variations in lighting,
pose and expression and only rough alignment.  Zhou et al.~\cite{ZhouEtAl2015}'s bi-channel approach used
a CNN to extract features from the low-resolution input face and then
mapped these using fully connected network layers to an intermediate
upsampled face image, which is linearly combined with a bicubicly
interpolated upsampling of the input face image to create a final
high-resolution output face.  In other work, Yu and
Porikli~\cite{Yu2016,Yu2017} used GANs for $8$ times super-resolution
of face images.  Their method provides visually appealing results, but
the resulting images can distort the identity of the person, which is
a critical issue for face recognition applications.  Cao et
al.~\cite{Cao2017} presented a deep reinforcement learning approach
that sequentially discovers attended patches followed by facial part
enhancement. Zhu et al.~\cite{Zhu2016} proposed a bi-network
architecture to solve super-resolution in a cascaded way. Each of
these methods is intended to produce visually pleasing face images and
do not consider face recognition accuracy.  They do not test their
methods on a face recognition task.

A recent paper by Ledig et al.~\cite{Ledig2017} that use a GAN for
general image super-resolution is similar in spirit to ours in the
sense that they also use the feature vector from a pre-trained CNN in
the loss function used to optimize their network.  In their case, in addition to reconstruction and adversarial losses they
use many feature maps from a VGG-19 network~\cite{Simonyan2014} trained on
ImageNet~\cite{DengEtAl2009} to compare the similarity of upsampled and
reference images (with any content) and achieve at most $4$ times super-resolution.  In our case we use the single
penultimate feature vector from a VGG Deep Face
network~\cite{Parkhi2015} to compare upsampled and reference face
images in order to train a face-specific super-resolution network that
maintains identity for $8$ times magnification.

To the best of our knowledge, our paper is the first to use a deep
neural network for face-specific super-resolution which is optimized
not for visual quality, but for face recognition accuracy.  We show
that our super-resolution algorithm improves over other
state-of-the-art super-resolution algorithms in terms of face
verification accuracy on both controlled face datasets (FRGC~\cite{Phillips2005}
and Multi-PIE~\cite{MultiPIE}) as well as an in-the-wild datasets
(LFW-a~\cite{Huang2007,Wolf2011} and MegaFace~\cite{Kemelmacher2016megaface}).

\hide{
1) Face verification is well studied in HR faces, for LR faces the question is what is the minimal resolution for a method -- give details about minimal resolution and what very low resolution mean and state the resolution of the faces we are interested in

2) Two types of methods for LR FR: (1) indirect methods - first step is SR (2) direct methods -- aim is to obtain a resolution-robust feature representation. We fall into indirect methods, but we do face verification rather than rec ognition

3) Talk about face SR methods: (1) methods that aim at most 4x magnifictaion [Kanade's limits of SR method as a kind of survey], (2) recent deep learning based methods aiming 8x magnification --- here I have to point out the exact face resolution these algorithms are tackling [oncel, porikli]

4) Recognition-based SR methods for LR FR.

5) The us eof recognition-based loss for superresolution

}

\section{Method}

\subsection{Notation}
We denote $\mathbf{x}_L^i\in\mathbb{R}^{N\times M}$ and $\mathbf{x}_H^i\in\mathbb{R}^{dN \times dM}$ as a pair of low-resolution and high-resolution (\ie $d$ times larger) versions of the $i$-th face image, function $\mathbf{G}: \mathbb{R}^{N\times M} \rightarrow \mathbb{R}^{dN\times dM}$ as a high-resolution image generator from low-resolution images, function $\mathbf{f}: \mathbb{R}^{dN\times dM} \rightarrow \mathbb{R}^D$ as a $D$-dim feature extractor from high-resolution images, $\|\cdot\|$ and $\|\cdot\|_F$ as the $\ell_2$ norm of a vector and the Frobenius norm of a matrix, respectively.

\subsection{Face Verification Problem Setup}
\noindent
{\bf Training:} We are provided with a set of $K$ pairs as well as their identities, \ie $\left\{\left(\mathbf{x}_L^i, \mathbf{x}_H^i, y_i\right)\right\}_{i=1}^K$, where $y_i\in\mathcal{Y}$ denotes the identity of the $i$-th image pair. We would like to learn face verification models based on such training data.

\noindent
{\bf Testing:} We are provided with a new pair of low-resolution (as probe) and high-resolution (as gallery) face images, and asked whether these two images share the same identity based on the learned models.

\subsection{Algorithm}

An overview of our proposed approach is shown in Figure~\ref{fig:overview}. We first super-resolve the given low-resolution face image using a deep convolutional network. Next, we extract features from the super-resolved image and a high-resolution gallery image using the VGG deep face network~\cite{Parkhi2015}. The similarity of the two images is decided based on the Euclidean distance between their feature vectors. Finally the verification is performed with a thresholding on the feature distance. 

\subsubsection{Training Objective}
We start our explanation from the objective function for training our model. 
Inspired by conventional methods, we propose optimizing the following objective function:
\begin{align}\label{eqn:obj}
\min_{\mathbf{G}, \mathbf{f}} \mathfrak{L}\left(\left\{\left(\mathbf{x}_L^i, \mathbf{x}_H^i, y_i\right)\right\}_{i=1}^K, \mathbf{G}, \mathbf{f}\right) + \lambda_1\Omega_1(\mathbf{f}) + \lambda_2\Omega_2(\mathbf{G}),
\end{align}
where $\mathfrak{L}$ denotes the loss function, $\Omega_1, \Omega_2$ denote two regularizers on $\mathbf{f}$ and $\mathbf{G}$ (\eg weight decay), respectively, and $\lambda_1\geq0, \lambda_2\geq0$ denote the predefined constants. In particular, we decompose the loss function as follows:
\begin{equation}
 \mathfrak{L} \stackrel{\mbox{def}}{=}\mathfrak{L}_f\left(\left\{\left(\mathbf{x}_H^i, y_i\right)\right\}_{i=1}^K, \mathbf{f}\right) + \mathfrak{L}_{recog} \\
\end{equation}
where $\mathfrak{L}_f$ denotes the classification loss (\eg least square) used in conventional recognition approaches for measuring the performance of $\mathbf{f}$, $\mathfrak{L}_{recog}$ denotes the {\em recognition loss} to measure the performance of $\mathbf{G}$ given $\mathbf{f}$
\begin{equation}\label{eqn:recog}
 \mathfrak{L}_{recog} = \sum_i{\omega_i \left \| \vec{f}(\mat{G}(\mat{x}_L^i))- \vec{f}(\mat{x}_H^i) \right \|}, 
\end{equation}
where $\omega_i\geq 0, \forall i$ denotes a weighting constant, in general, and in our current implementation we simply set $\omega_i=\frac{1}{K}$. Further investigation on the effect of varying $\omega_i$'s will be conducted in our future work.

Since this loss term computes the similarity of super-resolved and ground truth high-resolution faces, it can be used as an evaluation measure to assess the capacity of a method to preserve identity during super-resolution. 

\noindent
{\bf Discussion:} There are two alternative loss functions to recognition loss that are widely used in image super resolution or restoration~\cite{Zhao2017}.

\noindent
\underline{\em (1) Reconstruction Loss:} It measures the difference (in Euclidean space) between the reconstructed image from a low-resolution image and its corresponding high-resolution image, defined as follows:
\begin{equation}\label{eqn:rec}
\mathfrak{L}_{recon} = \frac{1}{K}\sum_i{\left \| \mat{G}(\mat{x}_L^i)-\mat{x}_H^i \right \|_F}.
\end{equation}
Minimizing this loss usually introduces a large amount of smoothing and averaging artifacts in reconstruction images that helps improve visual appearance but not verification accuracy necessarily.

\noindent
\underline{\em (2) Structural Similarity (SSIM) Loss:}
SSIM is a popular measure that accounts for humans perception of image quality. On a local patch of two images $\mat{x}$ and $\mat{y}$ around pixel $\vec{p}$, SSIM is computed as
\begin{equation}
SSIM(\mat{x}, \mat{y}, \vec{p}) = \frac{2\mu_x\mu_y+c_1}{\mu_x^2+\mu_y^2+c_1}\frac{2\sigma_{xy}+c_2}{\sigma_x^2+\sigma_y^2+c_2},
\end{equation}
where $\mu_x, \mu_y$ and $\sigma_x, \sigma_y$ denote the mean and variance, respectively, of the intensities in the local patch around pixel $\vec{p}$ in  $\mat{x}$ and $\mat{y}$ and  $\sigma_{xy}$ is the covariance of intensities in the two local patches. $c_1$ and $c_2$ are constant factors to stabilize the division with weak denominator. We divide the image into $h\times h$ grids and compute the mean of SSIM over all patches as the similarity between two images
\begin{equation}
r(\mat{x},\mat{y}) = \frac{1}{T}\sum_\vec{p}{SSIM(\mat{x}, \mat{y}, \vec{p})}
\end{equation} 
where $T$ is the total number of patches.
Consequently, our SSIM loss is formulated as
\begin{equation}\label{eqn:ssim}
\mathfrak{L}_{ssim} = \frac{1}{K}\sum_i{\left [ 1 - r\left(\mat{G}(\mat{x}_L^i), \mat{x}_H^i\right) \right ]}.
\end{equation}
Compared with reconstruction loss, minimizing SSIM loss helps recover the information at high frequency visually, but still unnecessarily improves the verification performance.

In our experiments we conduct comprehensive comparison on these three loss functions to demonstrate the correct usage of recognition loss for the task of face verification.

\subsubsection{Two-Stage Minimization}
To optimize Eq. \ref{eqn:obj} we propose using two-stage minimization technique. Precisely, we first learn the feature extraction function $\mathbf{f}$ {\em supervisedly} by optimizing
\begin{align}\label{eqn:obj_f}
\min_{\mathbf{f}} \mathfrak{L}_f\left(\left\{\left(\mathbf{x}_H^i, y_i\right)\right\}_{i=1}^K, \mathbf{f}\right) + \lambda_1\Omega_1(\mathbf{f}).
\end{align}
Then we learn the high-resolution image generator function $\mathbf{G}$ {\em unsupervisedly} based on the learned $\mathbf{f}$ by optimizing
\begin{align}\label{eqn:obj_G}
\min_{\mathbf{G}} \frac{1}{K}\sum_i{\left \| \vec{f}(\mat{G}(\mat{x}_L^i))- \vec{f}(\mat{x}_H^i) \right \|} + \lambda_2\Omega_2(\mathbf{G}).
\end{align}

%

\noindent
{\bf Two-Stage \vs End-to-End:} 
We implement end-to-end training algorithm as well, but find that the performance is much worse than our current two-stage minimization algorithm. We hypothesize that the end-to-end training involves many more parameters that need to be optimized, leading to overfitting on training data due to higher model complexity with respect to limited data samples. In contrast, our two-stage training strategy serves as regularization similar to the early stopping criterion used in deep learning.

\noindent
{\bf Network Architecture:} We use a similar architecture to Yu et al.~\cite{Yu2016} for our face super-resolution network (\ie function $\mathbf{G}$), except our super-resolution network is trained on gray-scale images. Our face recognition network (\ie function $\mathbf{f}$) has a VGG architecture with 19 layers as reported in~\cite{Parkhi2015}. 
Note that face descriptors from other deep networks for face recognition could be used instead of VGG, but this was chosen because the implementation of VGG deep face network is publicly available, and achieves near state-of-the-art performance on face recognition~\cite{Parkhi2015}.  \hide{what does it mean, so vague}

\subsubsection{Face Verification at Test Time}

After learning the super-resolution network, each super-resolved image is represented with its VGG face descriptor. The decision of whether a low-resolution face image $\vec{x}_L$ and a high-resolution gallery image $\vec{x}_H$  contain the same person is given based on the indicator function
\begin{equation}
I(\vec{x}_L, \vec{x}_H,\gamma) = \left\{\begin{matrix}
1 & \mbox{if} \, \left \| \vec{f}(\mat{G}(\mat{x}_L))- \vec{f}(\mat{x}_H) \right \|<\gamma, \\ 
0 & \mbox{otherwise}, 
\end{matrix}\right.
\end{equation}
where $\gamma$ is a threshold that can be determined using cross-validation.

\section{Experiments and Results}

\hide{ 

1) Algorithm parameters, information for reproducibility
2) datasets, how I obtained LR images, what are the datasets
3) Experimental setup for face verification - what is baseline, what is LR FV
4) Quantitative results: tables
visualization: Comparative Distance matrices + T-SNE
5) Qualitative results: visual outputs from both datasets

}

\subsection{Datasets and Experimental Setup}

We carried out two sets of experiments under controlled and uncontrolled, i.e. in the wild, settings. For controlled settings, we used Face  Recognition Grand Challenge (FRGC)~\cite{Phillips2005} and Multi-PIE~\cite{MultiPIE} datasets. For uncontrolled setting, we used an aligned version of the Labeled Faces in the Wild dataset~\cite{Huang2007}, called Labeled Faces in the Wild-a (LFW-a)~\cite{Wolf2011} and MegaFace dataset~\cite{Nech2017}.

\noindent
\textbf{FRGC:} The FRGC dataset contains frontal face images taken in a studio setting under two lighting conditions with only two facial expressions (smiling and neutral). We generated training and test splits, where we kept the identities in each set disjoint. The training set consisted of $20,000$ images from $409$ subjects and the test set consisted of $2,149$ images from $142$ subjects.

\noindent
\textbf{Multi-PIE:} The Multi-PIE dataset consists of face images of 337 subjects captured from various viewpoints and illumination conditions over multiple sessions. Our goal in using this set was to better evaluate the peformance of face verification under different facial poses and illumination conditions in a controlled setting. We use the three most frontal views (05\_1, 05\_0, 14\_0) and the four most frontal lighting conditions (06, 07, 08, 09) from each data collection session. We randomly generated training and test splits, where we kept the identities in each set disjoint. The training set consisted of $9,091$ images from $252$ subjects and the test set consisted of $3,000$ images from $85$ subjects. For both FRGC and Multi-PIE datasets, we carried out face alignment~\cite{Xiong2013}.

\noindent
\textbf{LFW-a:} This dataset consists of faces captured in an uncontrolled setting with several poses, lightings and expressions. We used the training and test splits as indicated in the LFW development benchmark, which also contains a set of image pairs to be tested in the verification task. The benchmark contains $9,525$ training images, $3,708$ test images and $1,000$ image pairs from the test set to be verified. 

\noindent
\textbf{MegaFace:} MegaFace consists of 4.7M images from 672K identities collected from Flickr users. We followed the experimental protocol described for face verification in MegaFace challenge 2. The training set is provioded along with facial landmark points. We performed face alignment using provided landmarks. We discarded the images with  resolution smaller than our high resolution images and the images with high registration error during alignment. As a result we used 2.8M images from MegaFace for training. Following the verification protocol, we used FaceScrub dataset~\cite{Ng2014} as our probe images during testing. FaceScrub comprises a total of 106,863 face images of male and female 530 celebrities. Negative pairs for verification are constructed using 10K distractor images provided by MegaFace challenge. Since facial landmarks are not provided for original high resolution images of FaceScrub and MegaFace distractor images, we carry out face alignment following~\cite{Tuzel2016b}.

\noindent
\textbf{Data preparation}

High-resolution images had $128\times128$ pixel resolution, where the faces occupied approximately $50\times50$ pixels area. We generated low-resolution images with $8$ times downsampling by following the approach in~\cite{Yang2014}. Namely, we filtered the high-resolution images with a Gaussian blur kernel $\sigma=2.4$ followed by downsampling. As a result, the low-resolution images contained a facial area of approximately $6\times6$ pixels. For the task of low-resolution face verification, we tested whether a given low-resolution probe image contains the same person as a given high-resolution gallery image. In the FRGC and Multi-PIE datasets, we considered verification of all test image pairs. For LFW-a dataset, we tested all $1,000$ image pairs in the benchmark, where each pair was tested two times by switching probe and gallery. For MegaFace dataset, we followed the verification protocol given in the MegaFace challenge.  More specifically, negative pairs consisted of all pairs between FaceScrub as probe images and MegaFace distractor dataset as gallery images. As for positive pairs, we tested each probe image from FaceScrub with the other images of the same identity as gallery. 

In terms of a baseline, we compare against face verification using the original high-resolution images for both gallery and probe (See Figure~\ref{fig:overview}).  In this case, our ``baseline'' algorithm provides an upper bound on the accuracy for low-resolution face recognition.  

\subsection{Algorithm Details}

We trained the VGG deep face network using the VGG face dataset~\cite{Parkhi2015} for $256\times256$ pixel gray-scale face images. The network outputs $4096$ dimensional feature vectors. Next, we trained the super-resolution network for each dataset separately by minimizing each of the loss terms stated in equations~\eqref{eqn:recog}, \eqref{eqn:rec} and \eqref{eqn:ssim}. We also analyzed how joint optimization of all the terms affects the verification performance by  minimizing weighted sum of all loss terms
\begin{equation}\label{eqn:joint}
\mat{\hat{G}} = \underset{\mat{G}}{argmin}\left ( {\mathfrak{L}_{recon}  + \alpha \mathfrak{L}_{ssim} + \beta \mathfrak{L}_{recog}  } \right ).
\end{equation}
 We set weights $\alpha$ and $\beta$ following a greedy search procedure. The code is implemented using the Caffe deep learning framework~\cite{Jia2014caffe}. Optimization was performed using the RMSProp algorithm~\cite{Hinton2012} with a learning rate of $0.001$ and a decay rate of $0.01$. The training took 3 days on a  NVIDIA GeForce GTX TITAN X GPU with Maxwell architecture and 12GB memory. Average running time for verification of a probe and gallery image pair is $0.1$ second. We used a patch size of $h=8$ for SSIM loss.

\subsection{Quantitative Results}

\begin{table}[t]
\centering 
\begin{tabular}{| l || c | c | c | c || c | c | c | c |} 
 \hline
\multirow{2}{*}{Method} & \multicolumn{4}{c||}{FRGC} &  \multicolumn{4}{c|}{Multi-PIE} \\  \cline{2-9}
& $\mathfrak{L}_{recog}$ & AUC  & PSNR & SSIM & $\mathfrak{L}_{recog}$ & AUC  & PSNR & SSIM \\
 \hline\hline
Bicubic	& 1.135 &	0.767 &	23.812 &	0.606 &	 1.170 &	0.850 &	18.923 &	0.443    \\ \hline
SRCNN~\cite{Dong2014} &	1.115	&	0.773 &	23.733& 	0.619 & 1.153 &	0.875 &	23.249 &	0.610\\ \hline
URDGN~\cite{Yu2016} &	1.025	& 0.780	 &17.738	& 0.512	& 1.143	& 0.896	& 23.845	& 0.632  \\ \hline
VDSR~\cite{Kim2016} &	1.088	& 0.796	& 24.794	& 0.646	& 1.192	& 0.685	& 13.366	& 0.285	\\ \hline
MZQ~\cite{Ma2010} &	0.909	& 0.806	& 25.287	& 0.758	& 1.170	& 0.850	& 18.923	& 0.443	 \\ \hline
Only $\mathfrak{L}_{recon}$ &	0.834	& 0.818	& \textbf{26.485}	& \textbf{0.797}	& 0.912	& 0.958	& \textbf{24.718}	& \textbf{0.724} \\ \hline
Only $\mathfrak{L}_{ssim}$ &	1.185	& 0.618	& 14.075	& 0.068	& 0.994	& 0.929	& 22.504	& 0.640\\ \hline
Only $\mathfrak{L}_{recog}$ &	0.794	& 0.831	& 15.730	& 0.247	& \textbf{0.879}	& 0.963	& 18.176	& 0.431 \\ \hline
Joint					&\textbf{0.788} & \textbf{0.833}	& 26.169	& 0.747	& 0.887	& \textbf{0.968}	& 24.428	& 0.684	 \\ \hline 
\end{tabular}
\caption{Quantitative results on FRGC and Multi-PIE, where baseline, (high-resolution face verification) AUC between original image pairs are computed as $0.851$ and  $0.998$ respectively. Best value for each column is shown in bold. }
\label{table:all_results}
\end{table}

\begin{table}[t]\footnotesize
\centering 
\begin{tabular}{| l || c | c | c | c || c | c | c | c |} 
 \hline
\multirow{2}{*}{Method} & \multicolumn{4}{c||}{LFW-a} &  \multicolumn{4}{c|}{MegaFace} \\  \cline{2-9}
& $\mathfrak{L}_{recog}$ & AUC  & PSNR & SSIM & $\mathfrak{L}_{recog}$ & AUC  & PSNR & SSIM \\
 \hline\hline
Bicubic	& 1.131 &	0.791 &	22.273 &	0.566 & 1.174	& 0.568	& 20.615	& 0.514 \\ \hline
SRCNN~\cite{Dong2014} &	1.067	 &	0.826	& 	22.833 &	0.601  & 1.126	& 0.656	& 21.177	& 0.546 \\ \hline
\small{URDGN~\cite{Yu2016} } & - & -& - & - & 1.107 &	0.730	& 16.804 &	0.401 \\ \hline
VDSR~\cite{Kim2016} & 1.074	& 0.845	 & \textbf{23.408} &	0.621 & 1.109 &	0.686	& 21.680	& 0.573 \\ \hline
MZQ~\cite{Ma2010} & 0.992	& 0.849	& 22.660	& 0.623 & 1.033	& 0.804	 & 21.496	& 0.604 \\ \hline
\small{Only $\mathfrak{L}_{recon}$ }& 1.018	& 0.850	& 22.655	&\textbf{ 0.625} & 0.981	& 0.848	&\textbf{ 22.674}	& \textbf{0.672}\\ \hline
\small{Only $\mathfrak{L}_{ssim}$ }& 1.159	& 0.673	& 13.417	& 0.171 & 1.253 & 0.406	& 11.305	& 0.071 \\ \hline
\small{Only $\mathfrak{L}_{recog}$ } & 0.974	& 0.883	& 16.600	& 0.336 & \textbf{ 0.900 }&	\textbf{0.891}	& 15.354	& 0.376 \\ \hline
Joint					& \textbf{0.963}	& \textbf{0.887}	& 22.055	& 0.537 & 0.949	& 0.864	& 21.218	& 0.555 \\ \hline 
\end{tabular}
\caption{Quantitative results on LFW-a and MegaFace datasets, where baseline, (high-resolution face verification) AUC between original image pairs are computed as $0.980$ and  $0.976$ respectively. Best value for each column is shown in bold. }
\label{table:wild_results}
\end{table}

\begin{figure}[t]
\begin{tabular}[t]{ccccccc}
Probe & 	Bicubic &  	$\mathfrak{L}_{recon}$ &	 $\mathfrak{L}_{ssim}$ & 	$\mathfrak{L}_{recog}$ &	 Joint & GT Probe\\
\includegraphics[width=0.13\textwidth]{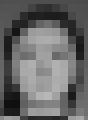}& 
\includegraphics[width=0.13\textwidth]{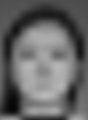}& 
\includegraphics[width=0.13\textwidth]{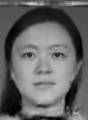}& 
\includegraphics[width=0.13\textwidth]{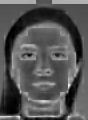}& 
\includegraphics[width=0.13\textwidth]{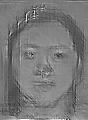}& 
\includegraphics[width=0.13\textwidth]{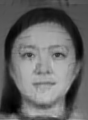}& 
\includegraphics[width=0.13\textwidth]{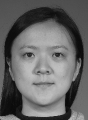}
\end{tabular}
\caption{Example probe images and super-resolution results on FRGC dataset. From left to right: low-resolution probe image, bicubic interpolation, super-resolution results with reconstruction, SSIM, recognition and joint losses, ground truth (GT) high-resolution probe image.}
\label{fig:frgc_results}
\end{figure}

We report quantitative results in Table~\ref{table:all_results} and Table~\ref{table:wild_results} for the experiments on controlled and uncontrolled datasets, respectively. In order to evaluate face verification performance, we computed a receiver operating characteristic (ROC) curve (plotting true positive versus false positive face verifications) by varying the threshold $\gamma$ and reported area under curve (AUC). Since an important contribution of our method is minimizing the distance of low-resolution and high-resolution images in feature space, we also report recognition loss $\mathfrak{L}_{recog}$ on the test set  as a means of quantifying identity preservation. For evaluating appearance quality, we report peak signal-to-noise ratio (PSNR) and SSIM. As seen in the tables, the network trained by minimizing recognition loss outperforms all other methods and the other individual loss functions in terms of face verification, although the others yield higher PSNR and SSIM values. Note that SSIM loss is computed on each image patch independently as opposed to the SSIM value that we compute for the whole image during evaluation. Low SSIM value using $\mathfrak{L}_{ssim}$ loss on FRGC is due to the fact that the images are all frontal and aligned, yielding filters learned independently for each facial patch (Please see block affects on visual results in Figure~\ref{fig:frgc_results}). 

\begin{figure}[t]
\centering
\begin{tabular}[t]{cc}
\includegraphics[width=2.0in]{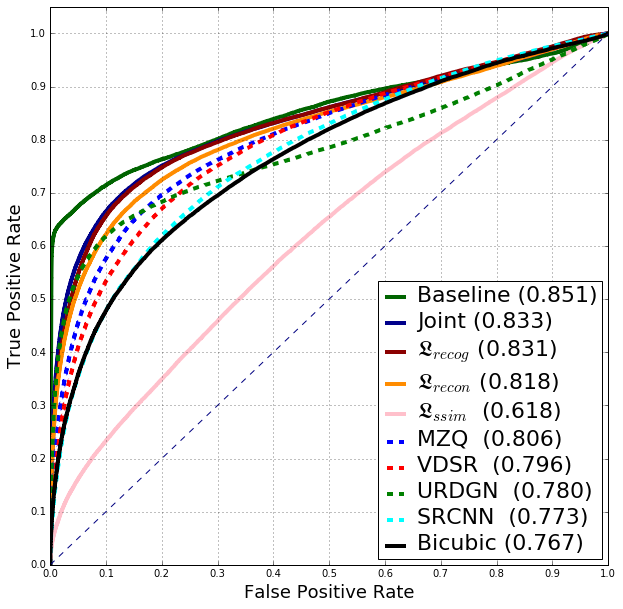} & \includegraphics[width=2.0in]{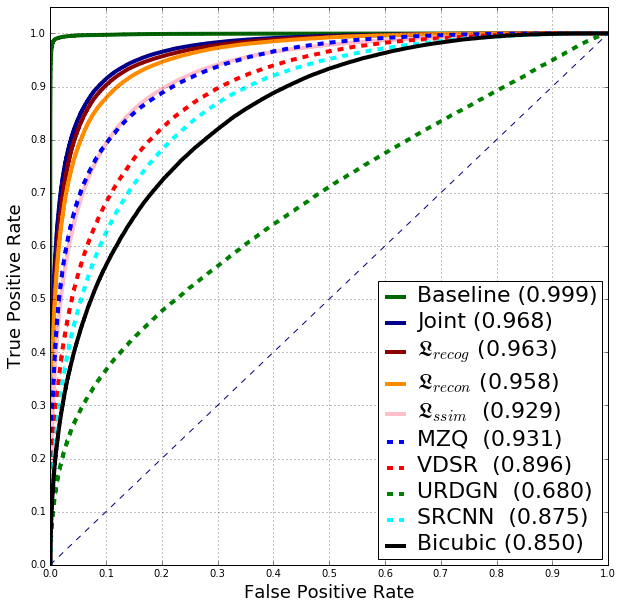} \\
(a) FRGC & (b) Multi-PIE \\
\includegraphics[width=2.0in]{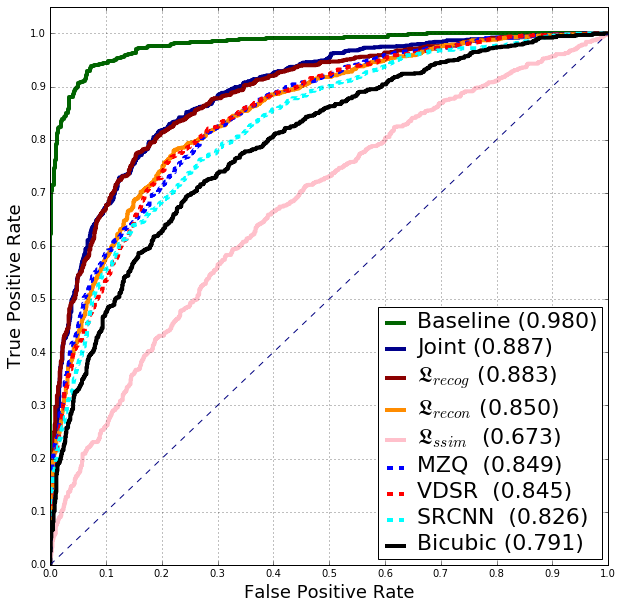} & \includegraphics[width=2.0in]{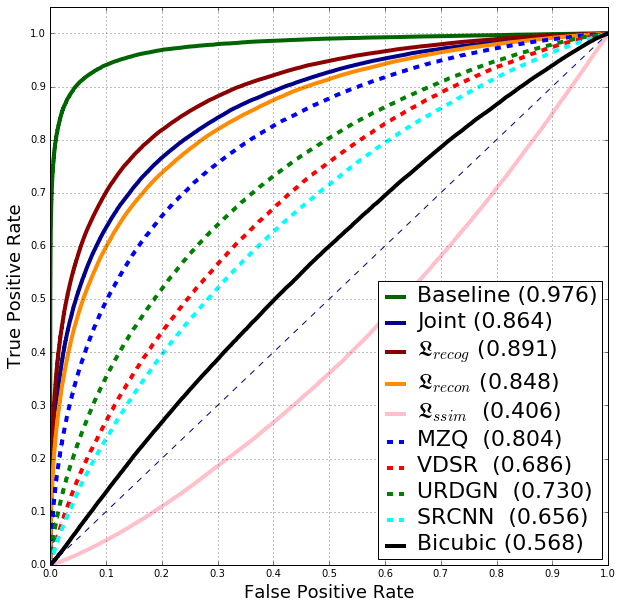}  \\
(c) LFW-a & (d) MegaFace \\
\end{tabular}
\caption{ROC curves for the results on (a) FRGC, (b) Multi-PIE, (c) LFW-a, and (d) MegaFace datasets. The numbers in parantheses indicate area under curve (AUC).}
\label{fig:all_roc}
\end{figure}

After a greedy procedure, we set weights of joint optimization for FRGC and Multi-PIE as $\alpha=1,000, \beta=300$ and for LFW-a and MegaFace as  $\alpha=10,000, \beta=3,000$. Note that the magnitude of $\mathfrak{L}_{recon}$ and $\mathfrak{L}_{ssim}$ is in the order of number of pixels in the high-resolution image and in the patch respectively, while the magnitude of $\mathfrak{L}_{recog}$ is in the order of feature vector dimension, which is $4096$. In Table~\ref{table:all_results} the results for joint optimization of all terms are comparable to the results with recognition loss. However, adjusting the weights of the terms in the joint optimization is a tedious process. Therefore, using only recognition loss is sufficient, if the final goal of the super-resolution is face identification. During greedy selection procedure, we also trained the network without SSIM loss, but the verification performance was slightly worse compared to using all three terms together. In Table~\ref{table:wild_results}, verification accuracy is comparable for joint loss and recognition loss on LFW-a, while on MegaFace dataset recognition loss outperforms with a difference of around $0.3$ in AUC. \hide{The results on MegaFace also shows the strength of the proposed method in a large-scale uncontrolled setting, where the performance of the proposed method is closer to the baseline compared to the results on other datasets.}\hide{Not sure this next sentence is helpful since recognition loss alone is all that is needed to match recognition loss: Hence, the use of SSIM loss in joint optimization was necessary to reach a comparable verification performance with the recognition-based loss.}

\begin{figure}[t]
\centering
\begin{tabular}[t]{cccc}
\includegraphics[width=0.19\textwidth]{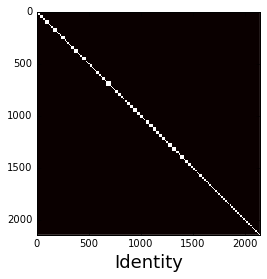}& 
\includegraphics[width=0.21\textwidth]{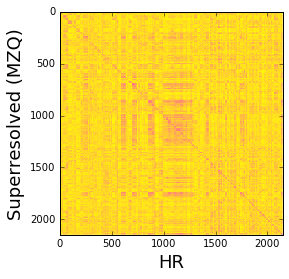} &
\includegraphics[width=0.21\textwidth]{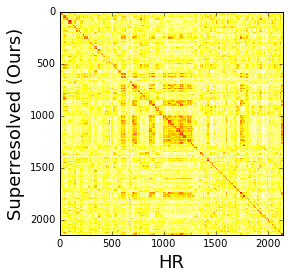}& 
\includegraphics[width=0.245\textwidth]{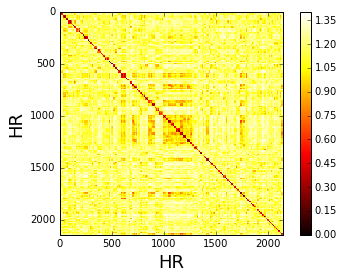}\\
(a) Labels & (b) MZQ method & (c) Our method & (d) Baseline \\
\end{tabular}
\caption{Colormap visualization of distance matrices: (a) facial identity matrix (white pixels indicate the images of the same person) and the distance matrices in feature domain between (b) super-resolved images with MZQ technique~\cite{Ma2010} and high-resolution images, (c) super-resolved images by our method and high-resolution images, and (d) among high-resolution image pairs, i.e. baseline.}
\label{fig:frgc_distMat}
\end{figure}

We compared the performance of our method with the state-of-the-art generic and face-specific super-resolution methods as reported in the table. For general object super-resolution methods, we compared with two deep learning-based methods: SRCNN by Dong et al.~\cite{Dong2014} and VDSR by Kim et al.~\cite{Kim2016}. Since both methods handle at most $4\times$ magnification, we performed $4\times$ magnification followed by $2\times$ magnification in order to achieve the same upsampling with our method. For face-specific super-resolution methods, we compared against URDGN by Yu et al.~\cite{Yu2016} and MZQ by Ma et al.~\cite{Ma2010}. We also performed experiments with structured face hallucination (SFH) technique~\cite{Yang2013}, but we omitted its results since it was not successful for most of the images for $8\times$ magnification due to its dependence on facial landmark detection. For the MZQ method we used the implementation of Yang et al.~\cite{Yang2013} and employed the same training set as ours for training except MegaFace dataset. Since MZQ is a dictionary-based approach and the implementation had memory constraints, for MegaFace dataset we randomly selected 100K images from the training set and used them for training MZQ method. For the other methods, we used the provided pre-trained models by the respective studies. URDGN method was trained on colored images with a different face alignment than ours. Therefore, we tested their method after aligning the test images according to their settings. Also, URDGN experiments were carried out on colored images, but final evaluation measures were computed on grayscale images. Note that all datasets consist of colored images except LFW-a, which has only grayscale images.

Figure~\ref{fig:all_roc} shows the ROC curves for all methods in each of the datasets. The difference between recognition and other losses is more visible in uncontrolled datasets, while recognition loss still outperforms the other losses  in both controlled setting datasets, FRGC and Multi-PIE. The super-resolution network trained with recognition loss has accuracy very close to the high-resolution (optimal) baseline for FRGC and Multi-PIE (around $0.03$ in terms of AUC), although for LFW-a and MegaFace there is a larger gap.  Compared to other state-of-the-art super-resolution algorithms, our network trained on $\mathfrak{L}_{recog}$ does significantly better.  The network trained on joint loss is slightly better than $\mathfrak{L}_{recog}$ for FRGC and LFW-a and slightly worse for Multi-PIE, at the expense of more time consuming training. Proposed method gives the best performance on large-scale MegaFace dataset. Our network trained on $\mathfrak{L}_{recon}$ is better in terms of AUC on all test sets than competing super-resolution methods, although not as good as when using $\mathfrak{L}_{recog}$ or joint loss.

Our main goal in this study is to obtain super-resolved images that are similar to the corresponding ground truth high-resolution (HR) images in the feature domain. Thus, we also provide a colormap visualization of the pairwise distance matrices in the feature domain for the FRGC dataset in Figure~\ref{fig:frgc_distMat}. As can be seen, the distance matrix of HR-to-HR (i.e. baseline) pairs is very similar to the distance matrix of super-resolved-to-HR pairs. Figure~\ref{fig:frgc_distMat}b also shows the distance matrix for MZQ algorithm~\cite{Ma2010} that gives the best performance among the comparative methods in terms of face verification. The difference between matching and unmatching pairs is more visible in the distance matrix obtained from our method. 

\setlength{\tabcolsep}{0.15em}

\setlength{\tabcolsep}{0.1em}
\begin{figure}[t]
\begin{tabular}[t]{cccccccc}
 \scriptsize Probe & Bicubic & \scriptsize MZQ~\cite{Ma2010} & \scriptsize VDSR~\cite{Kim2016} & \scriptsize SRCNN~\cite{Dong2014} & \tiny URDGN~\cite{Yu2016}  & Proposed & GT \normalsize\\ 
\includegraphics[width=0.115\textwidth]{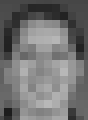}& 
\includegraphics[width=0.115\textwidth]{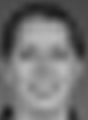}& 
\includegraphics[width=0.115\textwidth]{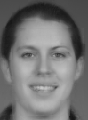}& 
\includegraphics[width=0.115\textwidth]{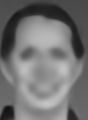}& 
\includegraphics[width=0.115\textwidth]{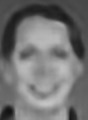}& 
\includegraphics[width=0.115\textwidth]{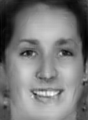}& 
\includegraphics[width=0.115\textwidth]{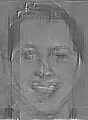}& 
\includegraphics[width=0.115\textwidth]{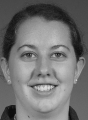}
\end{tabular}
\caption{Comparative results on FRGC dataset.}
\label{fig:frgc_comp}
\end{figure}

\subsection{Qualitative Results}

The goal of this study is not to obtain good looking super-resolved images, but rather to improve face verification accuracy for low-resolution images. We display visual results from our super-resolution network in order to elaborate what kind of features are retained and input to the VGG network for better face identification. 

Face verification was performed between a low-resolution probe image and high-resolution gallery image. Figure~\ref{fig:frgc_results} shows example low-resolution probe images along with the  super-resolved images from the minimization of various loss functions on the FRGC dataset. Reconstruction loss and SSIM loss smooth out facial details, while recognition-based loss can yield more details around important facial regions such as eyes and nose.

Figure~\ref{fig:frgc_comp} shows comparative super-resolution results from all tested methods on FRGC dataset. The column labeled ``Proposed'' is our super-resolution network trained using recognition loss only.  As can be seen face-specific super-resolution methods yield visually appealing results. However, their face verification performance is poor. 

\setlength{\tabcolsep}{0.1em}
\begin{figure*}[t]
\begin{tabular}[t]{cccccccc}
 Probe & Bicubic & \scriptsize MZQ~\cite{Ma2010} & \scriptsize VDSR~\cite{Kim2016} & \scriptsize SRCNN~\cite{Dong2014} & \tiny URDGN~\cite{Yu2016} & Proposed & GT \normalsize \\ 
\includegraphics[width=0.115\textwidth]{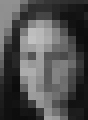}& 
\includegraphics[width=0.115\textwidth]{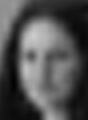}& 
\includegraphics[width=0.115\textwidth]{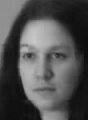}& 
\includegraphics[width=0.115\textwidth]{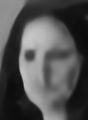}& 
\includegraphics[width=0.115\textwidth]{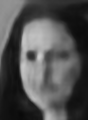}& 
\includegraphics[width=0.115\textwidth]{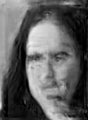}& 
\includegraphics[width=0.115\textwidth]{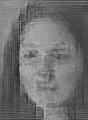}& 
\includegraphics[width=0.115\textwidth]{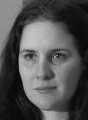}
\end{tabular}
\caption{Comparative results on Multi-PIE dataset.}
\label{fig:multiPIE_comp}
\end{figure*}

\setlength{\tabcolsep}{0.15em}
\begin{figure}[t]
\begin{tabular}[t]{ccccccc}
Probe & Bicubic & MZQ~\cite{Ma2010} & VDSR~\cite{Kim2016} & \scriptsize SRCNN~\cite{Dong2014} & Proposed & GT \\ 
\includegraphics[width=0.13\textwidth]{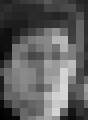}& 
\includegraphics[width=0.13\textwidth]{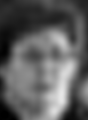}& 
\includegraphics[width=0.13\textwidth]{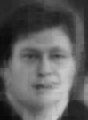}& 
\includegraphics[width=0.13\textwidth]{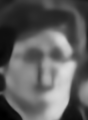}& 
\includegraphics[width=0.13\textwidth]{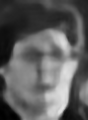}& 
\includegraphics[width=0.13\textwidth]{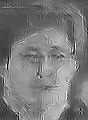}& 
\includegraphics[width=0.13\textwidth]{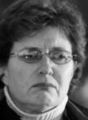}
\\ 
\includegraphics[width=0.13\textwidth]{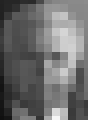}& 
\includegraphics[width=0.13\textwidth]{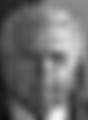}& 
\includegraphics[width=0.13\textwidth]{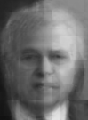}& 
\includegraphics[width=0.13\textwidth]{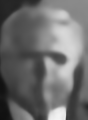}& 
\includegraphics[width=0.13\textwidth]{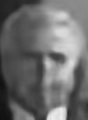}& 
\includegraphics[width=0.13\textwidth]{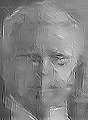}& 
\includegraphics[width=0.13\textwidth]{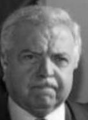}
\\ 
\end{tabular}
\caption{Comparative results on LFW-a dataset.}
\label{fig:lfwa_comp}
\end{figure}

\setlength{\tabcolsep}{0.1em}
\begin{figure}
\begin{tabular}[t]{cccccccc}
\includegraphics[width=0.115\textwidth]{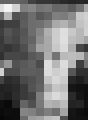}& 
\includegraphics[width=0.115\textwidth]{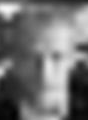}& 
\includegraphics[width=0.115\textwidth]{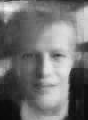}& 
\includegraphics[width=0.115\textwidth]{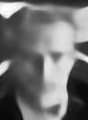}& 
\includegraphics[width=0.115\textwidth]{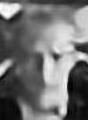}& 
\includegraphics[width=0.115\textwidth]{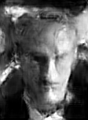}& 
\includegraphics[width=0.115\textwidth]{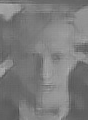}& 
\includegraphics[width=0.115\textwidth]{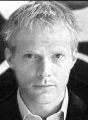}
\\ 
\includegraphics[width=0.115\textwidth]{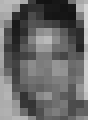}& 
\includegraphics[width=0.115\textwidth]{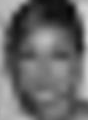}& 
\includegraphics[width=0.115\textwidth]{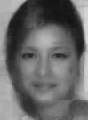}& 
\includegraphics[width=0.115\textwidth]{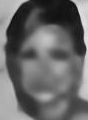}& 
\includegraphics[width=0.115\textwidth]{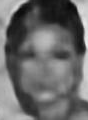}& 
\includegraphics[width=0.115\textwidth]{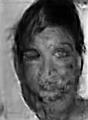}& 
\includegraphics[width=0.115\textwidth]{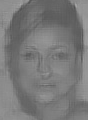}& 
\includegraphics[width=0.115\textwidth]{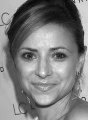}
\\ 
\end{tabular}
\caption{Comparative results on Megaface dataset.}
\label{fig:mf_comp}
\end{figure}

Figure~\ref{fig:multiPIE_comp}  shows visual results using various methods on the Multi-PIE dataset. The proposed method is not affected from uneven brightness and shadows on the face that occur due to lighting direction as seen in the figure. Comparative results on LFW-a and MegaFace datasets are displayed in Figures~\ref{fig:lfwa_comp} and~\ref{fig:mf_comp} respectively.

An interesting observation about faces upsampled using our proposed super-resolution network is that the lighting effects are largely removed, which is very useful if face recognition accuracy is the goal, although it is not what should happen if the goal is to create a high-resolution face image that looks like the input low-resolution face when downsampled.  \hide{The first two rows of }Figure~\ref{fig:lfwa_comp} and Figure~\ref{fig:mf_comp} illustrate this point the best. In Figure~\ref{fig:lfwa_comp}, the input faces have strong lighting on the right side of the image, but our proposed super-resolved faces are much more evenly lit. Similar effects are visible on the results in Figure~\ref{fig:mf_comp}. Removing the effects of lighting is well-known to improve face recognition accuracy.

\section{Conclusion and Discussion}

We presented a super-resolution-based method for verification of very low-resolution faces. Our method exploited a VGG network for feature extraction. We trained a deep neural network for $8$ times super-resolution of face images by minimizing the distances of high-resolution and super-resolved images of the same person in terms of their face descriptors computed by the VGG Deep Face network. The results on controlled and uncontrolled settings showed that the presented method provides better verification accuracy compared to conventional super-resolution techniques. This work demonstrates that generating visually appealing super-resolved images is not necessary if the final goal is improving face recognition accuracy.  Instead, the super-resolution network should directly optimize the distance to the desired face descriptor.

In this work, our aim was to learn a high-resolution image generator given a feature extractor, assuming that the feature extractor already performs well in high-resolution face verification. In other words, we trained the generator to find the best mapping function that will transform a given low-resolution image to its high-resolution version in feature space. Thus, feature extractor and generator were trained in alternating steps rather than an end-to-end fashion. Moreover, simultaneous learning of feature extractor and generator is a cumbersome task as it will require learning of a larger number of parameters.\hide{ unmatching image pairs in addition to matching image pairs.}

Using video inputs rather than single images is an important future extension of our work. Also, we would like to incorporate more constraints to our loss function to decrease the feature distance between face images with different identities.


\clearpage

\bibliographystyle{splncs}
\bibliography{refs}
\end{document}